\title{Mixed integer programming formulation of unsupervised learning}
\author{Arturo Berrones-Santos\\
{\small Universidad Aut\'onoma de Nuevo Le\'on }\\
{\small Facultad de Ingenier\' \i a Mec\'anica y El\'ectrica}\\
{\small Posgrado en Ingenier\' \i a de Sistemas}\\
{\small Facultad de Ciencias F\' \i sico Matem\'aticas}\\
{\small Posgrado en Ciencias con Orientaci\'on en Matem\'aticas}\\
{\small AP 126, Cd. Universitaria, San Nicol\'as de 
los Garza, NL 66450, M\'exico}\\
{\small arturo.berronessn@uanl.edu.mx}\\ 
}

\documentclass{article}
\usepackage{graphicx}
\usepackage{amsmath}
\usepackage[ruled]{algorithm}
\usepackage{algorithmic}

\begin{document}
\maketitle

\begin{abstract}
A novel formulation and training procedure for full Boltzmann machines in terms of
a mixed binary quadratic feasibility problem is given. As a proof of concept,
the theory is analytically and numerically tested on XOR patterns.

Keywords: Boltzamnn machines; Mixed integer programming; Unsupervised learning.
\end{abstract}

\section{Introduction}
\label{intro}
A central open question in machine learning is the effective handling of
unlabeled data \cite{anips,bengio}. The construction of balanced representative datasets for 
supervised machine learning for the most part still requires a very close and time 
consuming human direction, so the development of efficient
learning from data algorithms in an unsupervised fashion is a very active area of research \cite{anips,bengio}.
A general framework to deal with unlabeled data is the Boltzmann machine paradigm,
in which is attempted to learn a probability distribution for the patterns in the data
without any previous identification of input and output variables.
In its most general setups however, the training of Blotzmann machines is 
computationally intractable \cite{bengio,pr,arxiv}.
In this contribution is established a relation, which to the best of my knowledge was previously unknown,
between Mixed Integer Programing (MIP) and the full Boltzmann machine in binary variables.
Is hoped that this novel formulation opens the road to more efficient 
learning algorithms by taking advantage of the great variety of techniques available for MIP.

\section{Full Boltzmann machine with data as constraints}
Consider a network of units with binary state space. Each unit depends on all the others by a logistic-type response function,
\begin{eqnarray} \label{responses}
x_i =round \quad \left [\frac{1}{1+
\exp\left (-\sum_{j \neq i} q_{j,i}x_j - b_i \right)}\right] \equiv f_i,
\end{eqnarray}
where the ``round'' indicates the nearest integer function, the $q$'s are pairwise interactions between units
and the $b$'s are shift parameters.
As will later be clear, 
the proposed model supports both supervised and unsupervised learning and leads to a full Boltzmann machine in its classical sense.

\begin{figure}[!htb]
  \centering
      \includegraphics[width=0.5\textwidth]{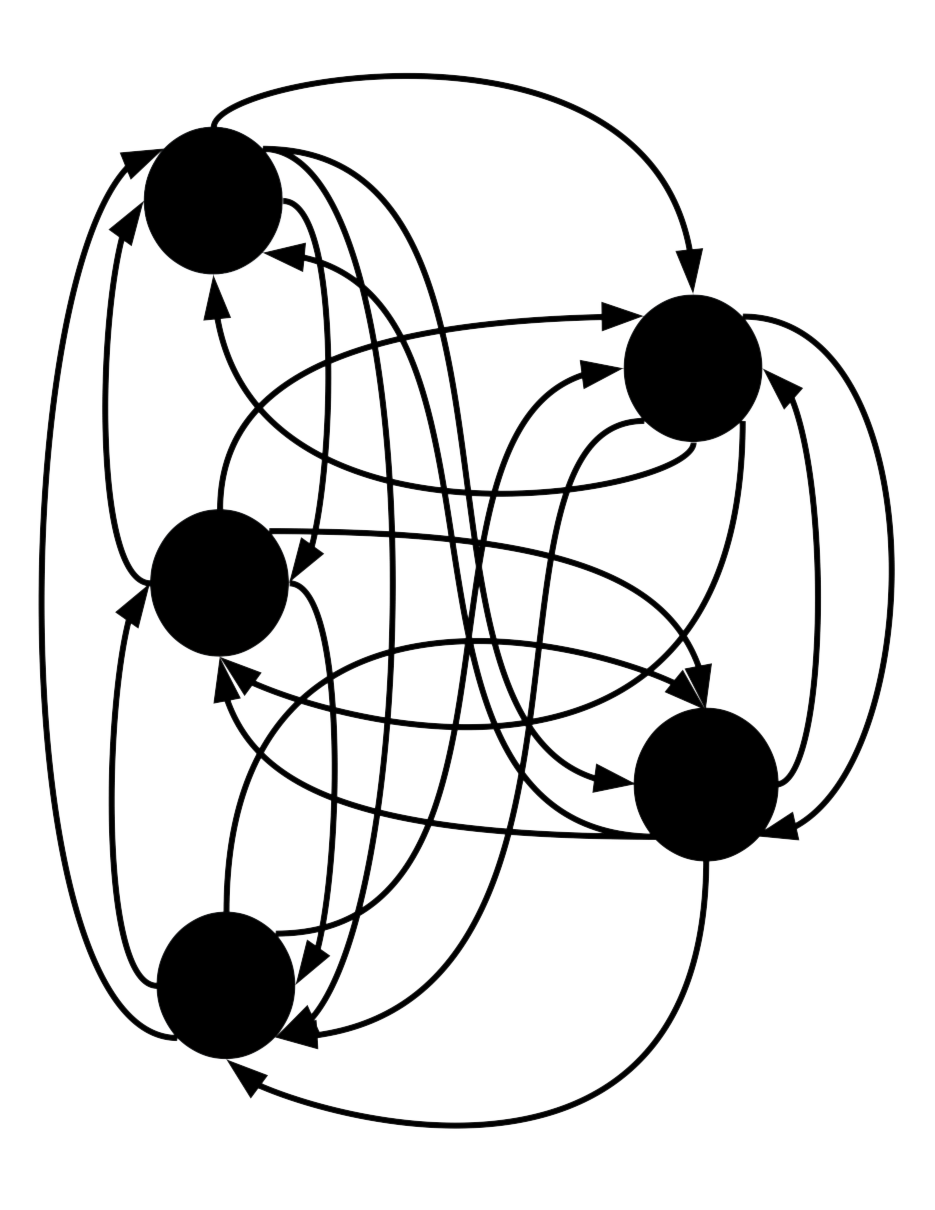}
  \caption{Full Boltzmann machine with five units.}
  \label{Fig1}
\end{figure}

Suppose a data set of $D$ visible binary vectors, $\{ \vec{v}_d \}$, $d = 1, 2,..., D$
with $I$ components each and a collection of $M$ hidden units $u_m$, $m=1,..., M$.
The total number of units in the system is $N = I + M$.
If the connectivity and shift parameters are given,
each sample fixes the binary vector $\vec{x}_d = \{ \vec{v}_d , \vec{u}_d\}$ and therefore
the data set imposes the following $ND$ constraints:
\begin{eqnarray}\label{constraints}
(-1)^{v_{d,i}} \left (\sum_{j \neq i}^{I} q_{i,j}v_{d,j}  + b_i\right ) \quad \leq 0 , \\ \nonumber
(-1)^{u_{d,m}} \left ( \sum_{j \neq m}^{M} q_{m,j}u_{d,j} + b_m \right ) \quad \leq 0 , \\ \nonumber
\quad d = 1, 2,..., D; i = 1, 2, ..., I; m = 1, 2, ..., M.
\end{eqnarray}
A posterior distribution for the parameters $P(\{q_{i,j}, b_i\} | D)$, can be constructed by 
the maximum entropy principle, which gives the less biased distribution that is 
consistent with a set of constraints \cite{jaynes,mep}.
This is done by the minimization of the Lagrangian
\begin{eqnarray}\label{lagrangian}
\mathcal{L} = \int P \ln P d\vec{w} + \sum_{r = 1}^{ND} \lambda_{r} \left < constraint(r) \right >,
\end{eqnarray}
where the brackets represent average under the posterior, $\vec{w}$ is a vector that contains 
the connectivity and shift parameters and 
the $\lambda$'s are positive Lagrange multipliers.
Due to the linearity of the system of inequalities (\ref{constraints}), the average of the constraints
under $P$ with fixed unit values is simply given by the same set of inequalities with the 
coefficients $q_{j,i}$'s and $b_i$'s substituted by their averages $\left < q_{j,i} \right >$'s, $\left < b_i \right >$'s.
The maximum entropy distribution for the parameters is therefore
given by 
\begin{eqnarray}\label{maxent}
P(\{q_{i,j}, b_i\} | D) = \frac{1}{Z} \exp \left [ - \sum_{\{d,i\}} \lambda_{\{d,i\}}\left [ (-1)^{x_{d,i}} \left (\sum_{j \neq i}^{N} q_{i,j}x_{d,j} + b_i \right ) \right ] \right ],
\end{eqnarray}
where $Z$ is a normalization factor.
So due to the linearity of the constraints, $P$ is a tractable (i. e. an easy to sample) product of
independent two parameter exponential distributions:
\begin{eqnarray}\label{maxent2}
P(\{q_{i,j}, b_i\} | D) = P(\vec{w} | D) = \prod_{n=1}^{N} \alpha_n e^{-\alpha_n (w_n - \beta_n)},
\end{eqnarray}
where $\left < w_n \right > = \frac{1}{\alpha_n} + \beta_n$ \cite{handbook}.
Therefore a necessary and sufficient condition for the existence of the above distribution is the existence of the averages
$\left < w_n \right >$, which is determined by the satisfaction of the inequalities (\ref{constraints}).

The representation of the posterior by its two parameter exponential form Eq. (\ref{maxent2})
gives a codification of the training data in terms of a tractable distribution for the parameters
that in conjunction with Eq. (\ref{responses}) is in fact a distribution for new unlabeled binary strings
of data. For fully connected topologies, this is what is usually understood by 
an equilibrium distribution of a full Boltzmann machine \cite{bengio}.

\begin{figure}[!htb]
  \centering
      \includegraphics[width=0.9\textwidth]{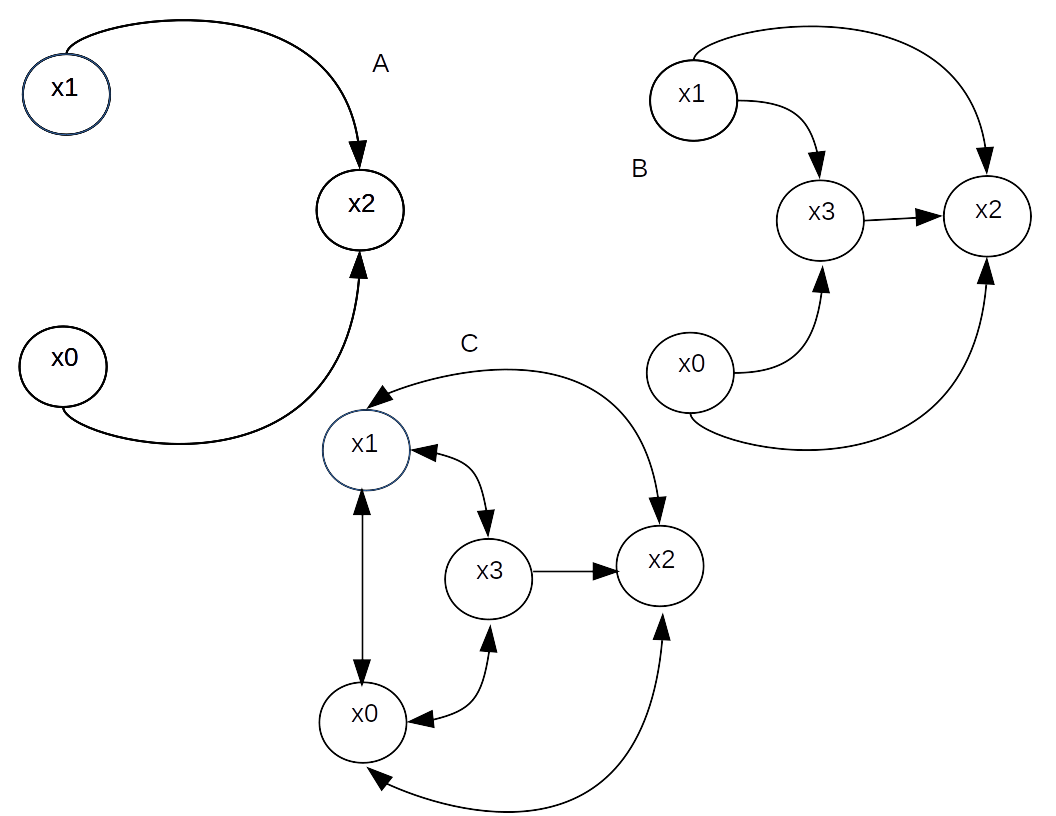}
  \caption{Three different architectures for the XOR problem.}
  \label{Fig2}
\end{figure}

\subsection{Illustrative example 1: Supervised XOR} \label{sxor}
The theoretical soundness of the proposed approach is now shown through the XOR logical table,
$D = \{ (0,0,0), (1,0,1), (0,1,1), (1,1,0) \}$.
Let's consider first a restricted architecture
with only two directed arcs that connect two inputs with an output unit, as 
represented in Figure \ref{Fig2}A.
The inequalities (\ref{constraints}) in this case read,
\begin{eqnarray}\label{inxor}
b_2 \leq 0, \quad - q_{0,2} - b_2 \leq 0, \\ \nonumber
- q_{1,2} - b_2 \leq 0, \quad q_{0,2} + q_{1,2} + b_2 \leq 0.
\end{eqnarray}
There are no values for $b_2$, $q_{0,2}$ and $q_{1,2}$ that satisfy all the inequalities.
This is reflected in the maximum entropy distribution,
\begin{eqnarray}
P = \frac{1}{Z} e^{-\lambda_1 (b_2)} e^{-\lambda_2 (- q_{0,2} - b_2)} e^{-\lambda_3 (- q_{1,2} - b_2) } e^{-\lambda_4 (q_{0,2} + q_{1,2} + b_2)}
\end{eqnarray}
which to be a properly normalized product of two-parameter exponential distributions must satisfy the contradictory
conditions $b_2 < 0$, $q_{0,2} > 0$, $q_{1,2} > 0$, $|b_2| < q_{0,2}$, $|b_2| < q_{1,2}$, $|b_2| > q_{0,2} + q_{1,2}$.
A valid model is however attainable by the addition of a single hidden unit. Consider the architecture represented
in Figure \ref{Fig2}B. This leads to a two stage constraint satisfaction problem. The first stage
is given by the data evaluated on the visible units,
\begin{eqnarray}
q_{3,2} f_3(0,0) + b_2 \leq 0, \\ \nonumber
- q_{0,2} - q_{3,2} f_3 (1,0) - b_2 \leq 0, \\ \nonumber
- q_{1,2} - q_{3,2} f_3 (0,1) - b_2 \leq 0, \\ \nonumber
q_{0,2} + q_{1,2} + q_{3,2} f_3 (1,1) + b_2 \leq 0,
\end{eqnarray}
for which solutions certainly exist. Take for instance, 
\begin{eqnarray}
|b_2| < q_{0,2}, \quad |b_2| < q_{1,2}, \quad |b_2 + q_{3,2}| > q_{0,2} + q_{1,2}, \\ \nonumber
f_3(0,0) = f_3(1,0) = f_3(0,1) = 0, \quad f_3(1,1) = 1.
\end{eqnarray}
The second stage is consequently given by,
\begin{eqnarray}
b_3 \leq 0, \\ \nonumber
q_{0,3} + b_3 \leq 0, \\ \nonumber
q_{1,3} + b_3 \leq 0, \\ \nonumber
- q_{0,3} - q_{1,3} - b_3 \leq 0,
\end{eqnarray}
for which solutions exist under the conditions $b_3 < -C$,
$|b_3| > |q_{0,3}|$, $|b_3| > |q_{1,3}|$, $|b_3| < |q_{0,3} + q_{1,3}|$, where $C$ is a positive constant.
Therefore, the maximum entropy distribution for the parameters of the
model represented in the Figure \ref{Fig2}B exists. Equivalently, this result shows that
the classical XOR supervised learning problem can be solved by the 
proposed MIP feasibility formulation.

\subsection{Illustrative example 2: Unsupervised XOR} \label{usxor}
A model capable of unsupervised learning is sketched in Figure \ref{Fig2}C.
The system of inequalities should be now extended to consider inputs to
nodes $x_0$ and $x_1$,
\begin{eqnarray}
q_{3,0} f_3(0,0) + b_0 \leq 0, \\ \nonumber
- q_{2,0} - q_{3,0} f_3 (1,0) - b_0 \leq 0, \\ \nonumber
q_{1,0} + q_{2,0} + q_{3,0} f_3 (0,1) + b_0 \leq 0, \\ \nonumber
- q_{3,0} f_3 (1,1) - q_{1,0} - b_0 \leq 0, \\ \nonumber
q_{3,1} f_3(0,0) + b_1 \leq 0, \\ \nonumber
q_{0,1} + q_{2,1} + q_{3,1} f_3 (1,0) + b_1 \leq 0, \\ \nonumber
- q_{2,1} - q_{3,1} f_3 (0,1) - b_1 \leq 0, \\ \nonumber
- q_{0,1} - q_{3,1} f_3 (1,1) - b_1 \leq 0,
\end{eqnarray}
which has indeed solutions, as discussed in the following section.

\section{Sampling from the posterior distribution} \label{pxor}
The equilibrium 
posterior distribution of patterns can be sampled by 
taking an arbitrary solution of the MIP feasibility problem 
and using it to define the averages 
$\left < w_n \right > = \frac{1}{\alpha_n} + \beta_n$.
The standard deviation of each two-parameter exponential distribution is given 
by $\sigma_n = \frac{1}{\alpha_n}$, which can be at first instance assigned to some positive
value related to the constant $C$.
If $y$ is an uniform random deviate in the interval $[0,1]$, then
$w_n = - \frac{1}{\alpha_n} \ln (1-y) + \beta_n$ is a deviate 
from the two-parameter exponential distribution associated to $w_n$.
In this way, the vector of visible units can be sampled in
a computation time that is quadratic in the number of total (visible and invisible) units.

\begin{algorithm}
\caption{(Pseudo-code for sampling from the maximum entropy posterior.)} \label{alg1}
\begin{algorithmic}[1]
\STATE Initialize: $\beta_n$ from a solution of the MIP feasibility problem and
$\frac{1}{\alpha_n} \leftarrow \epsilon C$ ($\epsilon$ arbitrary positive real number).
\STATE Asign value to $size$ (desired number of samples).
\STATE Generate $y_{\tau}$, $(\tau = 1,..., size \times N)$ uniform and independent random deviates in the $[0,1]$ interval.
\FOR{s = 1 \TO size} 
     \STATE {$w_{n,s} = - \frac{1}{\alpha_{n,s}} \ln (1-y) + \beta_{n,s}$, $n = 1, ... , N$} 
     \STATE Generate $\vec{x}_s$ by inserting $\vec{w}_s$ in Eq. (\ref{responses})
\ENDFOR
\end{algorithmic}
\end{algorithm}

The step $6$ of the algorithm above is made by starting with an inital random binary vector
$\vec{x}$ at each $s$. The self-consistent system Eq. (\ref{responses}) is then iterated. No more than
$10$ iterations are needed to achieve convergence.

The sampling procedure Algorithm \ref{alg1} is now shown through the XOR example.
Take an arbitrary solution of the MIP feasibility problem, say
\begin{eqnarray}
f_3(0,0) = f_3(1,0) = f_3(0,1) = 0, f_3(1,1) = 1, \\ \nonumber
-b_0 = -b_1 = -b_2 = -b_3 = C, \\ \nonumber
q_{0,3} = q_{1,3} = \frac{3}{4} C,  \\ \nonumber
q_{2,0} = q_{0,2} = q_{1,2} = q_{2,1} = -q_{0,1} = -q_{1,0} = 2C, \\ \nonumber
q_{3,0} = q_{3,1} = - q_{3,2} = 4C.
\end{eqnarray}
Due to the rounding operator in Eq. (\ref{responses}), any $C > 0$ can work.
In the following experiments the value $C=100$ is used with sample sizes of $1500$.
Some of the samples drawn for each $C$ are shown. 

\begin{eqnarray} \nonumber
(a): \quad \frac{1}{\alpha_n} = 0.1C \quad \forall \quad n: \\ \nonumber
[0 1 1],
[0 0 0],
[1 1 0],
[1 0 1],
[0 0 0],
[0 1 1],
[0 0 0], \\ \nonumber
[0 0 0],
[1 1 0], 
[1 1 0],
[0 0 0],
[0 1 1],
[0 0 0],
[0 0 0],
[1 1 0]...
\end{eqnarray}

\begin{eqnarray} \nonumber
(b): \quad \frac{1}{\alpha_n} = 0.5C \quad \forall \quad n: \\ \nonumber
[0 1 1],
[0 0 0],
[0 0 0],
[1 1 1],
[1 1 0],
[0 0 0],
[0 0 0], \\ \nonumber
[0 0 0],
[0 0 0],
[0 0 0],
[1 1 1],
[0 0 0],
[1 0 1],
[0 0 0],
[0 1 1]...
\end{eqnarray}

\begin{eqnarray} \nonumber
(c): \quad \frac{1}{\alpha_n} = 2.0C \quad \forall \quad n: \\ \nonumber
[1 0 0],
[0 0 1],
[1 1 1],
[0 1 1],
[0 0 0],
[0 0 0],
[1 0 1], \\ \nonumber
[0 0 0],
[0 0 1], 
[0 0 0],
[0 0 0],
[0 0 1],
[0 0 0],
[1 0 0],
[1 0 1]...
\end{eqnarray}

The resulting ratios of XOR patterns relative to
non-XOR patterns over the entire $1500$ samples for each case are,
$(a): 1$, $(b): 0.9$ and $(c): 0.6$

\section{Discussion}
In the author's view, this paper presents a formalism that 
has the potential not only to give more efficient 
learning algorithms but to improve the understanding of the learning from
data itself. Particularly, datasets explicitly constrain the parameters of
the learning model by a set of feasiblity mixed binary inequalities.
For fixed binary values, the system is linear and continuous. For fixed model
parameters, it's a linear constraint satisfaction problem in binary variables. 
The author together with collaborators is now working in different ways to
exploit these structures in order to scale the framework to solve realistic large scale
unsupervised learning problems.
In such problems, a measure proportional to the number of 
satisfied constraints might be used to gide the learning procedure and to assign sensible 
values to the $\frac{1}{\alpha_n}$ hyperparameter.

\section*{acknowledgements}
The author acknowledge partial financial support from UANL and CONACyT.

 \section*{Conflict of interest}

 The author declares that he have no conflict of interest.

\end{document}